\documentclass[runningheads]{llncs}

\usepackage{eccv}
\usepackage{eccvabbrv}

\usepackage{amsmath}
\usepackage{booktabs}
\usepackage{graphicx}
\usepackage{placeins}
\usepackage[accsupp]{axessibility}
\usepackage{xcolor}
\usepackage{hyperref}
\usepackage{orcidlink}
\hypersetup{hidelinks}

\newif\ifadvisorversion
\advisorversionfalse
\ifadvisorversion
  \colorlet{reviewcolor}{blue}
\else
  \colorlet{reviewcolor}{black}
\fi
\newenvironment{reviewaddition}{\color{reviewcolor}}{}

\begin{document}

\title{Gloss-Free Representation Learning for Cross-Dataset Sign Spotting}
\titlerunning{Gloss-Free Representation Learning for Sign Spotting}

\author{\textcolor{reviewcolor}{O\u{g}uz Akif T\"ufekcio\u{g}lu}\inst{1}\orcidlink{0009-0001-1578-8190} \and
\textcolor{reviewcolor}{Ezgi Ekin}\inst{1}\orcidlink{0009-0006-6008-4308} \and
\textcolor{reviewcolor}{Mustafa Kaan \c{C}evik}\inst{1}\orcidlink{0009-0000-0092-2660} \and
\textcolor{reviewcolor}{Hacer Yalim Keles}\inst{1}\orcidlink{0000-0002-1671-4126}}
\authorrunning{\textcolor{reviewcolor}{O. A. T\"ufekcio\u{g}lu et al.}}
\institute{\textcolor{reviewcolor}{Hacettepe University, Ankara, T\"urkiye}\\
\textcolor{reviewcolor}{\email{\{oguztufekcioglu,ezgiekin21,kaancevik21,hacerkeles\}@hacettepe.edu.tr}}}

\maketitle

\begin{abstract}
Sign-language research for resource-constrained languages is often limited by
the cost of dense linguistic labels, including glosses, temporal boundaries,
and sign order. Broadcast news provides a practical alternative by pairing
continuous signing with spoken-language transcripts, but this supervision is
weak because text and signing are only loosely aligned. Morphologically rich
languages such as Turkish add a further difficulty, since the same lexical
meaning can appear in many inflected forms, while some derived forms should
remain separate. We study whether weak transcript-based supervision can pretrain a reusable sign encoder in this morphologically rich setting, where inadequate text normalization can fragment pseudo-gloss targets and weaken representation learning. Unlike prior pseudo-gloss pipelines designed mainly to improve translation, we test whether the pretrained encoder transfers as a reusable representation for cross-dataset sign spotting. We pretrain on TSL-News, a Turkish broadcast corpus collected for this work, using pseudo-gloss labels derived from transcripts rather than manually annotated glosses. We compare two pseudo-gloss construction strategies, namely rule-based morphological lemmatization and constrained LLM-assisted normalization over a fixed vocabulary. We evaluate the learned representations through cross-dataset sign spotting on a new TSL Spotting Benchmark built from the TSL Dictionary corpus. The LLM-assisted encoder raises top-5 temporal localization mean IoU from 0.235 with raw spatial features to 0.465, with 56.2\% of examples reaching an IoU of at least 0.50. A frequency analysis further suggests that localization quality is not mainly driven by memorization of frequent pseudo-gloss labels. In a downstream translation check, the same pretraining improves BLEU-4 from 9.60 to 11.04 and ROUGE from 23.48 to 27.43. These results show that loosely aligned broadcast data can provide effective weak supervision for learning sign representations that capture both lexical content and temporal structure.

\end{abstract}

\section{Introduction}
Sign language translation (SLT)~\cite{decoster2023machine,shahin2024rulebased,bragg2019sign}, sign language recognition ~\cite{koller2020quantitative}, and sign spotting~\cite{bragg2019sign} all rely on visual representations that preserve both lexical content and temporal structure. However, the annotations commonly used to train such models are costly to obtain. Gloss labels require linguistic expertise, temporal boundaries require frame-level or segment-level annotation, and sign order may differ from the corresponding spoken-language sentence. These requirements are especially restrictive for resource-limited sign languages, where parallel corpora and expert gloss annotations are scarce.

Broadcast news data offer a practical source of continuous signing paired with spoken or written language translations. However, this supervision is weak: the text and signing are only loosely aligned, and they may differ in lexical choice, word order, and level of detail. These issues are more difficult in a morphologically rich spoken language such as Turkish, where many surface forms can refer to the same lexical unit, while some derived forms should remain distinct. The question addressed in this paper is therefore not whether a new spotting or translation architecture can be designed, but whether weak text supervision can be used to pretrain a reusable sign encoder. We study this question in the pretraining stage of a Turkish Sign2GPT-style pipeline~\cite{wong2024sign2gpt}, treating it as a gloss-free representation learning problem. The goal is to test whether sentence-level pseudo-gloss labels derived from Turkish text can shape a video encoder into a representation that transfers beyond the original broadcast data. We evaluate this transfer through sign spotting: given an isolated dictionary sign as a visual template, the model must localize the corresponding sign inside a sentence-level example video. Rather than evaluating only on the pseudo-gloss labels used during training, we test the learned representation on the TSL Dictionary corpus (TSLD), using isolated signs as templates and sentence videos as localization targets.

The evaluation uses normalized cross-correlation (NCC) between hidden-state sequences. Given an isolated TSLD sign video and a continuous TSLD sentence video, the isolated-sign representation is slid over the continuous representation, and the highest-scoring temporal windows are compared with human temporal annotations. This setup directly tests the representation-learning question. If the pretrained encoder has learned reusable sign structure, an isolated dictionary sign should match the corresponding sign inside a sentence video, although neither video comes from the broadcast corpus used for pretraining.

The results support this hypothesis. The encoder pretrained with the constrained LLM-assisted pseudo-gloss strategy improves NCC-based temporal localization over the unadapted spatial-feature baseline. Mean IoU increases from 0.124 to 0.271 at top-1, from 0.192 to 0.403 at top-3, and from 0.235 to 0.465 at top-5 on TSL Spotting Benchmark (TSL-SB). The same strategy also improves both vocabulary coverage and localization accuracy compared with the rule-based morphological lemmatization baseline. These gains are achieved under weak supervision. During pretraining, the model receives only sentence-level pseudo-gloss labels derived from Turkish text, without exact sign labels or temporal boundary annotations.

\paragraph{Contributions.}
Four contributions are made. First, we repurpose a Sign2GPT-style pseudo-gloss pretraining pipeline~\cite{wong2024sign2gpt} -- originally proposed to supply weak supervision for gloss-free translation -- to ask a question this line of work has not addressed: whether the resulting representation is itself strong enough to support cross-dataset sign spotting, a task central to both sign language recognition and translation. This reframes an existing weak-supervision pretraining paradigm as a representation-learning question, tested outside both its original translation objective and its training corpus. Second, we design two Turkish-specific pseudo-gloss construction strategies, moving from rule-based morphological lemmatization to a constrained LLM-assisted lexical normalization, to address the lexical variation introduced by Turkish's agglutinative morphology. Third, we document TSL-News and TSL-SB as broadcast pretraining and cross-dataset spotting resources for this setting. Fourth, we introduce a multi-angle evaluation protocol -- combining isolated-template NCC matching, temporal IoU, direct pseudo-gloss score localization, and all-vocabulary local NCC -- that isolates representation quality from confounds such as vocabulary coverage or classifier bias, complemented by a downstream translation check.

\section{Related Work}

\paragraph{Sign spotting and dictionary-based localization.}
Sign spotting asks whether and where a target sign appears in continuous,
co-articulated signing. Early modern spotting systems emphasize the same
practical difficulty faced here: isolated dictionary productions are useful
queries, but they differ from continuous broadcast signing in speed, context,
and signer behavior. Watch, Read and Lookup learns spotting embeddings from
sparse localized labels, subtitles, and isolated dictionary examples using a
multiple-instance contrastive objective, explicitly treating subtitles and
dictionaries as weak supervisors rather than dense gloss annotations
\cite{momeni2020watch}. Large-vocabulary continuous sign language recognition
(CSLR) from spoken-language supervision
extends this direction by using retrieval-style objectives, pseudo-label
cleaning, synonym aggregation, and dense evaluation with temporal sign
intervals on the BBC-Oxford British Sign Language (BOBSL) broadcast dataset
\cite{raude2024tale}. Recent segmentation work instead predicts temporal
boundaries directly with begin-inside-outside (BIO) labels, transformer
temporal modeling, hand-pose features, and connectionist temporal
classification (CTC) sequence constraints \cite{low2025handson}. These methods
show that spotting quality depends not only on a classifier, but also on the
representation used for cross-domain matching and on the temporal proposal
mechanism. The present work follows the dictionary-query view of spotting, but
uses it as an evaluation protocol: isolated TSLD signs are templates,
continuous TSLD examples are search videos, and NCC over hidden states tests whether weak
broadcast pretraining has learned reusable lexical-temporal structure.

\paragraph{Gloss-free and weakly supervised sign representations.}
Gloss annotations remain expensive because they require linguistic expertise,
sign-order decisions, and often temporal alignment. Gloss-free SLT therefore
tries to avoid a manually annotated gloss bottleneck. Gloss-Free Sign Language
Translation with Visual-Language Pretraining (GFSLT-VLP) aligns visual and
textual representations through visual-language pretraining and masked language
modeling without gloss supervision \cite{zhou2023gfsltvlp}, while
Sign2GPT uses automatically extracted pseudo-glosses to pretrain a visual
encoder before connecting it to a large language model \cite{wong2024sign2gpt}.
Other representation-learning work is also relevant: learnt contrastive
concept (LCC) embeddings use spoken-language word embeddings to regularize sign
embeddings and enable
automatic temporal localization \cite{wong2023lcc}, and SignRep learns
sign-specific self-supervised red-green-blue (RGB) representations that
transfer to recognition, dictionary retrieval, and translation
\cite{wong2025signrep}. These studies
support the broader premise that useful sign representations can be learned
without dense manual gloss boundaries. Our setting is narrower and more
diagnostic: the pseudo-gloss labels are Turkish text-derived sentence-level
bags, and success is measured by cross-dataset temporal localization rather
than by translation scores or isolated recognition accuracy.

\paragraph{Temporal alignment and post-processing.}
Weak supervision in continuous signing also requires careful temporal handling.
CSLR models commonly use sequence losses or alignment constraints to learn from
video-level gloss order without frame boundaries, and motion-aware encoders
such as CorrNet show that local hand and face trajectories are central to
continuous sign representations \cite{hu2023corrnet}. After visual scoring,
post-processing often determines whether weak predictions become usable
annotations: subtitle windows are padded or shifted, still frames are trimmed,
repeated high-confidence detections are merged, score thresholds suppress
isolated spikes, and synonym or subtitle alignment is used to reduce lexical
noise. Gloss Alignment uses spoken-language embeddings to reassign spotted
glosses to neighboring subtitles when subtitle timing and signing order are
misaligned \cite{walsh2023glossalignment}; large-language-model disambiguation
similarly reranks dictionary-based spotting candidates with sentence-level
linguistic context after visual matching \cite{low2025llmspotting}. In this paper, the
post-processing is deliberately simple--fixed trimming, NCC peak selection, and
top-$k$ temporal IoU--so that the experiment primarily probes representation
transfer. The failure cases therefore connect directly to the literature:
stronger proposal generation, context-aware reranking, and text-gloss
realignment are natural next steps once the encoder has shown cross-dataset
spotting signal.

\section{Method}

\subsection{Gloss-Free Weak Pseudo-Gloss Pretraining}

\begin{figure}[t]
  \centering
  \includegraphics[width=\linewidth]{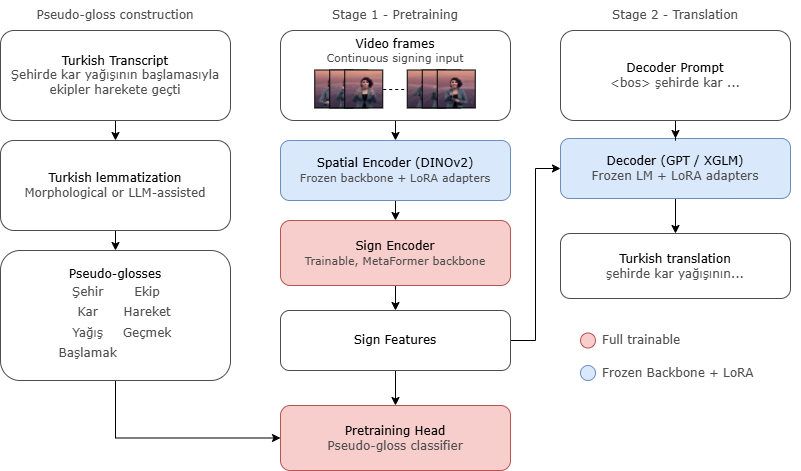}
  \caption{Overview of the pretraining and translation pipeline. Turkish text is converted into pseudo-glosses via lemmatization; video frames pass through a frozen DINOv2 encoder and a trainable sign encoder, whose output feeds both the pseudo-gloss pretraining head and, for context, a frozen GPT/XGLM decoder (both LoRA-adapted).}
  
\label{fig:full_pipeline}
\end{figure}

Let $V = \{x_t\}_{t=1}^{T}$ be a sign-language video and let $G(V) \subseteq \mathcal{G}$ be the set of pseudo-gloss labels extracted from the paired Turkish text. These labels are sentence-level bags of words rather than sign-language glosses with temporal boundaries; no manually annotated sign glosses are used during pretraining. The stage-1 model maps the video into temporal hidden states
\begin{equation}
H = E_\theta(V) \in \mathbb{R}^{T' \times d}.
\end{equation}
Here, $x_t$ is the input frame at time $t$, $T$ is the input video length, $\mathcal{G}$ is the pseudo-gloss vocabulary, $E_\theta$ is the trainable video encoder with parameters $\theta$, $H$ is the sequence of hidden states, $T'$ is the output sequence length after temporal downsampling, and $d$ is the hidden feature dimension. Note that $t$ indexes the input video at frame resolution ($t = 1, \dots, T$), whereas the downsampled hidden-state sequence $H$ is indexed separately below. In the current model setting, $E_\theta$ uses DINOv2~\cite{oquab2023dinov2}, a self-supervised vision-transformer backbone, for frame features with lightweight adaptation, followed by a MetaFormer~\cite{yu2022metaformer} temporal encoder for token mixing with local attention and downsampling. The output hidden states are passed to a prototype head whose class prototypes are initialized from Turkish fastText~\cite{bojanowski2017enriching} subword word embeddings. For class $g$ at position $t' \in \{1, \dots, T'\}$ in the downsampled sequence, the head computes a cosine similarity $s_{t',g}$ between a projected hidden state and the word prototype. The temporal class score combines class and time normalization,
\begin{equation}
p_g = \sum_{t'} \operatorname{softmax}_{g}\left(s_{t',g}/\tau_c\right) \operatorname{softmax}_{t'}\left(s_{t',g}/\tau_t\right),
\end{equation}
Here, $p_g$ is the sentence-level score for pseudo-gloss $g$, $\operatorname{softmax}_g$ normalizes similarities across pseudo-gloss classes, and $\operatorname{softmax}_{t'}$ normalizes across the $T'$ downsampled time steps. The trainable temperatures $\tau_c$ and $\tau_t$ control the sharpness of the class and time distributions in the TSL-News pretraining setting. The model is trained with binary cross entropy (BCE) over the sentence-level pseudo-gloss set:
\begin{equation}
\mathcal{L}_{\mathrm{pg}} = \operatorname{BCE}\left(\{p_g\}_{g \in \mathcal{G}}, \{\mathbb{1}[g \in G(V)]\}_{g \in \mathcal{G}}\right).
\end{equation}
Here, $\mathcal{L}_{\mathrm{pg}}$ is the pseudo-gloss pretraining loss and $\mathbb{1}[g \in G(V)]$ is an indicator target that equals 1 when pseudo-gloss $g$ appears in the text-derived label set for video $V$, and 0 otherwise. Because no temporal target is supplied, any temporal structure in $H$ must emerge from video dynamics, the encoder inductive bias, and the pressure to explain the weak pseudo-gloss set. Figure~\ref{fig:full_pipeline} summarizes the full pipeline, including pseudo-gloss construction, stage-1 pretraining, and the downstream translation stage that reuses the pretrained visual encoder.

\begin{reviewaddition}
Technical details on Sign2GPT~\cite{wong2024sign2gpt} pretraining are provided in Appendix~\ref{sec:supp_sign2gpt_comparison}.
\end{reviewaddition} 

\subsection{Turkish Pseudo-Gloss Construction}

Turkish Sign Language broadcast pretraining requires converting Turkish sentence translations into weak lexical targets. This is not a trivial word-filtering step because Turkish is highly agglutinative: inflectional suffixes can mark case, possession, number, person, tense, and modality, while derivational suffixes can create related but lexically different words~\cite{oflazer1994spelling,oflazer1994tagging}. A surface-word vocabulary therefore fragments the same lexical concept across many forms, but overly aggressive normalization can also merge targets that should remain distinct for signing. We therefore compare two Turkish-specific pseudo-gloss construction strategies, both of which produce sentence-level lexical sets rather than manual sign glosses.

\paragraph{Morphology-lemma strategy.} The first strategy uses rule-based Turkish morphological analysis, following the finite-state/two-level morphology tradition for agglutinative Turkish~\cite{oflazer1994spelling,oflazer1994tagging}. For each Turkish surface token, the analyzer returns possible morphological parses; each parse is decomposed into a structured representation, and the root morpheme of the first inflectional group is treated as a lemma candidate. Since this stage does not perform sentence-level morphological disambiguation, multiple analyses of the same token are resolved by selecting the root that appears most often among the returned parses. Tokens with no root analysis are discarded, one-character lemmas are removed, and repeated lemmas inside the same sentence are collapsed because the pretraining objective supervises pseudo-gloss presence rather than token count. This strategy mainly removes inflectional variation, so forms such as case-, possessive-, plural-, tense-, or agreement-marked words contribute to a shared lemma target. Compared with a raw, unlemmatized surface-token baseline (not manually annotated glosses) of approximately 16K targets, this yields a pseudo-gloss vocabulary of 4802 classes. Its limitation is that analyzer ambiguity, named entities, derivational morphology, and broadcast phrasing can still create lexical targets that do not correspond cleanly to signed units.

\paragraph{LLM lexical-rule strategy.} The second strategy uses an LLM as a constrained lexical normalizer rather than as a free gloss generator. The normalization rules select content-bearing Turkish units -- nouns, proper names, verbs, adjectives, adverbs, pronouns, and numbers -- while suppressing function words such as conjunctions and question particles. Inflectional morphology is removed; verbs are normalized toward an active infinitive or base form; spelling, diacritic, and proper-name variants are canonicalized; and derivational suffixes are preserved when they change the lexical meaning. Passive and causative verb variants are pruned when a simpler base verb is already available, reducing redundant supervision from closely related verbal forms.

To keep the output closed-vocabulary, translation tokens are then matched against the allowed pseudo-gloss inventory using exact lexical matches, root/\allowbreak prefix matches, and fuzzy string candidates; the LLM may only choose from these candidates or reject the token. This turns the LLM step into a rule-constrained disambiguation and normalization stage rather than an unbounded annotation source. The resulting pseudo-gloss vocabulary contains 6,539 classes, 1737 more than the morphology-lemma vocabulary. Thus, the two alternatives differ not only in size, but also in how they trade off coverage, derivational specificity, and lexical consistency. The resulting TSLD coverage difference is reported in Table~\ref{tab:construction}, where Overlap (\%) is computed relative to a fixed 2004 pool of TSLD sign-word entries considered matchable under either method, not relative to each method's own pseudo-gloss vocabulary size.

Additional implementation details for the LLM lexical-rule pseudo-gloss construction, including the prompt structure, mapping statistics, and training hyperparameters, are provided in the Supplementary Material.

\FloatBarrier

\section{Datasets and Cross-Dataset Evaluation Protocol}

\subsection{Datasets}

\paragraph{TSL-News.}
We introduce TSL-News, a new Turkish news broadcast corpus, to support weak-supervision-based techniques in the sign language domain (Table~\ref{tab:tsl_news}).
It contains television (TV) news videos from 2021--2023 with
Turkish transcripts, but no manual gloss labels, sign-order labels, or temporal
sign boundaries. The corpus therefore matches the limited-resource setting of
this paper: supervision is available only at the sentence level, while the
model must learn temporally useful sign representations from noisy
spoken-language text.

\begin{table}[!h]
\centering
\small
\caption{TSL-News broadcast pretraining corpus.}
\label{tab:tsl_news}
\setlength{\tabcolsep}{4pt}
\begin{tabular}{lp{3.2cm}p{5.4cm}}
\toprule
Property & Value & Implication \\
\midrule
Source & TV news broadcasts, 2021--2023 & Broadcast signing, not laboratory recording \\
Sentence segments & 13,378 & 11,507 train / 802 validation / 1,069 test \\
Duration / frames & 21+ hours / $\sim$3.6M frames & Large continuous video corpus \\
Average segment & $\sim$6 s, $\sim$10 words & Sentence-level supervision \\
Signers & 3 & Limited signer diversity \\
Annotations & Turkish transcript only & No boundaries, gloss labels, or sign order \\
\bottomrule
\end{tabular}
\end{table}

\paragraph{TSLD.}
The Turkish Sign Language dictionary corpus (TSLD) is used for transfer
evaluation. TSLD is based on the publicly available TID Dictionary corpus ~\cite{makaroglu2017guncel}; the local crawl contains 2004 dictionary headword entries with isolated sign videos and example or meaning videos. It provides sign-word dictionary entries, isolated sign videos, and
sentence-level example videos. For this work, TSL-SB is prepared by matching
TSLD sign-word entries or single-token synonym variants to the
TSL-News pseudo-gloss vocabularies (the matching procedure is detailed in Sec.~\ref{sec:benchmark_construction}), then manually temporally annotating sign occurrences
in example videos.
This separates training from evaluation: the encoder is pretrained on
broadcast TSL-News, but spotting is measured on dictionary and example videos
from TSLD.

\subsection{Benchmark Construction}
\label{sec:benchmark_construction}

The evaluation links two resources. TSL-News supplies the broadcast training
distribution and pseudo-gloss vocabularies. TSLD supplies sign-word dictionary
entries, isolated sign videos, and continuous example sentence videos. An
overlap vocabulary is constructed by scanning TSL-News pseudo-gloss tokens and
matching TSLD sign-word dictionary entries or single-token synonym variants, after
removing a conservative list of Turkish stopwords. Multiword phrase matches are
intentionally avoided to reduce noisy overlap from frequent function words.

\begin{table}[!h]
\centering
\small
\caption{TSLD overlap construction statistics with TSL-News for the two
pseudo-gloss construction methods. Vocabulary sizes identify the corresponding
weak-supervision settings. Overlap (\%) is normalized by the fixed pool of
2004 matchable TSLD sign-word entries, not by the pseudo-gloss vocabulary
size.}
\label{tab:construction}
\setlength{\tabcolsep}{8pt}
\begin{tabular}{lrrr}
\toprule
Pseudo-gloss method & Vocab Size & Overlap & Overlap (\%) \\
\midrule
Morphology-lemma & 4802 & 1060 & 52.89 \\
LLM lexical-rule & 6539 & 1398 & 69.76 \\
\bottomrule
\end{tabular}
\end{table}

TSL-SB contains 596 annotated sign words and 1842
sentence-level temporal
annotations, of which 1817 fall inside the LLM lexical-rule
vocabulary and 1159 inside the morphology-lemma vocabulary,
with a shared intersection of 1137 used for the controlled comparison in
Sec.~\ref{Sec:5.1} (the underlying label export contains 2127 segment-level intervals,
whose role is detailed in Sec.~\ref{Sec:5.3}). Together with the
isolated sign videos used to build template
representations, the benchmark contains 4363 videos. Because the model output
layer is defined by its pseudo-gloss dictionary, each evaluation filters
examples whose target sign word is absent from the active vocabulary.
\subsection{NCC Localization}

For an annotated TSLD example, let $V_{\mathrm{sent}}$ be the continuous
sentence video and $V_{\mathrm{iso}}$ be the isolated dictionary video. Then
$R = E_\theta(V_{\mathrm{sent}})$ is the continuous sentence representation and
$Q = E_\theta(V_{\mathrm{iso}})$ is the isolated-sign representation. The
isolated sequence $Q$ is slid over $R$, and normalized cross-correlation
(NCC)~\cite{lewis1995fast} is computed at offset $u$:
\begin{equation}
    \operatorname{NCC}(u) =
    \frac{1}{|Q|}
    \sum_{i=1}^{|Q|}
    \left\langle
    \frac{Q_i-\mu_Q}{\sigma_Q},
    \frac{R_{u+i}-\mu_{R,u}}{\sigma_{R,u}}
    \right\rangle .
\end{equation}
Here, $u$ is the candidate start offset in the continuous sequence, $|Q|$ is
the isolated-template length, $Q_i$ is the $i$th isolated hidden state, and
$R_{u+i}$ is the aligned continuous hidden state. The terms $\mu_Q$ and
$\sigma_Q$ are the mean and standard deviation of the isolated
template, computed per feature dimension over the $|Q|$
template steps, while
$\mu_{R,u}$ and $\sigma_{R,u}$ are the corresponding statistics of the
continuous window starting at $u$. The inner product compares the standardized
template and continuous features. The top-$k$ NCC peaks define candidate
temporal windows of length $|Q|$. Each window is compared against the human annotation using
interval IoU, and the best IoU among the first $k$ ranked candidates is
reported for $k \in \{1,3,5\}$. All main runs trim 0.5 seconds from both ends
of the continuous search region and 0.2 seconds from both ends of isolated
signs. This trimming removes non-signing margins because idle frames can
dominate NCC matching if they are left in the search/template sequences.

\FloatBarrier

\section{Experiments}

\subsection{Main NCC Results}
\label{Sec:5.1}
Table~\ref{tab:ncc_main} reports NCC localization performance. The
spatial-feature baseline uses raw pretrained spatial features before stage-1
pseudo-gloss pretraining, evaluated on the same example set as
the LLM lexical-rule vocabulary (hence the identical $N$ and Skip values). The trained LLM lexical-rule model improves top-1 mean IoU by
0.147 absolute over this baseline and nearly doubles top-5 mean IoU
(0.465 vs. 0.235). It also improves top-5 IoU@0.50 from 23.9\% to 56.2\%.

\begin{table}[!h]
\centering
\small
\caption{NCC temporal localization. $N$ is the number of evaluated examples,
Skip is the number filtered out by vocabulary mismatch, and Top-5@0.50 is the
percentage of examples whose best top-5 candidate reaches at least 0.50 IoU.}
\label{tab:ncc_main}
\setlength{\tabcolsep}{3.8pt}
\begin{tabular}{lrrrrrr}
\toprule
Representation & N & Skip & Top-1 & Top-3 & Top-5 & Top-5@0.50 \\
\midrule
Spatial features & 1817 & 25 & 0.124 & 0.192 & 0.235 & 23.9\% \\
Morphology-lemma encoder & 1159 & 683 & 0.173 & 0.293 & 0.363 & 42.8\% \\
LLM lexical-rule encoder & 1817 & 25 & \textbf{0.271} & \textbf{0.403} & \textbf{0.465} & \textbf{56.2\%} \\
\bottomrule
\end{tabular}
\end{table}

\paragraph{Pseudo-gloss Construction and Coverage.}
The LLM lexical-rule strategy increases the matched TSLD sign vocabulary from 1,060 to 1,398 signs. It also reduces the number of benchmark examples skipped due to vocabulary mismatch from 683 to 25. In the full coverage-oriented comparison in Table~\ref{tab:ncc_main}, the LLM lexical-rule encoder is evaluated on 1817 examples, while the morphology-lemma encoder is evaluated on 1159 examples; therefore, these rows should be read as both a representation and coverage comparison. To control for the different evaluated example counts, we additionally compute scores on the 1137 annotated example groups covered by both vocabularies. On this shared subset, the morphology-lemma encoder reaches 0.175/0.297/0.368 top-1/top-3/top-5 IoU, while the LLM lexical-rule encoder reaches 0.271/0.412/0.473, corresponding to gains of 0.096/0.115/0.105.
\begin{figure}[!h]
\centering
\includegraphics[width=0.98\linewidth]{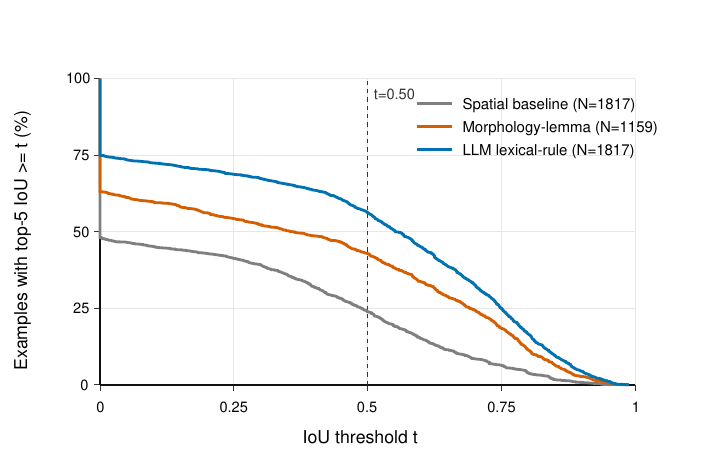}
\caption{Top-5 IoU survival curves for NCC localization. Each curve uses its
own evaluated set: spatial features ($N{=}1817$), morphology-lemma
($N{=}1159$), and LLM lexical-rule ($N{=}1817$). The dashed line marks
$t{=}0.50$, corresponding to the Top-5@0.50 column in
Table~\ref{tab:ncc_main}.}
\label{fig:ncc_iou_survival}
\end{figure}

Figure~\ref{fig:ncc_iou_survival} shows the full top-5 IoU distribution behind
the averages in Table~\ref{tab:ncc_main}. The curves make two effects visible. Where each curve meets
the vertical axis ($t{=}0$) shows how many examples were localized at all:
the spatial baseline starts near 48\%, meaning the sign is completely missed
(top-5 IoU~$=0$) in about half of the examples, while the LLM lexical-rule
encoder starts near 75\%, cutting the complete-miss rate to about one quarter.
The curves also shift to the right, meaning that when a sign is found, it is
found more precisely: the median top-5 IoU rises from 0.000 for the spatial
baseline and 0.358 for the morphology-lemma encoder to 0.553 for the LLM
lexical-rule encoder.

\begin{figure}[!h]
\centering
\includegraphics[width=0.98\linewidth]{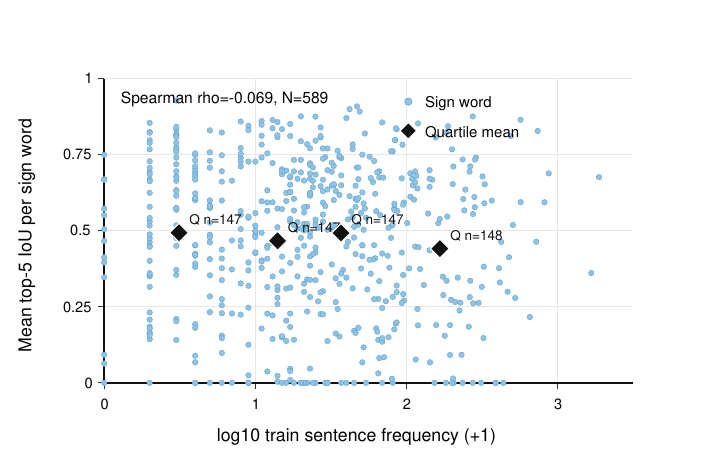}
\caption{TSL-News train frequency versus TSL-SB localization for the LLM
lexical-rule encoder. Each point is one matched sign word ($N{=}589$); the
$x$-axis is \(\log_{10}(\mathrm{frequency}{+}1)\), where frequency counts train
sentences whose pseudo-gloss set contains the target. Black diamonds show
frequency-quartile means. Spearman \(\rho{=}-0.069\).}
\label{fig:iou_frequency_scatter}
\end{figure}

Figure~\ref{fig:iou_frequency_scatter} relates localization quality to the
number of TSL-News train sentences in which each target pseudo-gloss appears.
The association is weak (\(\rho{=}-0.069\)), and the quartile means remain
close even for low-frequency signs, suggesting that the LLM lexical-rule
encoder is not simply memorizing frequent pseudo-gloss labels.
Two caveats apply. By construction, the benchmark contains
only signs that appear at least once in the pretraining text, so truly unseen
signs are not tested. Within this matched range, however, localization quality
does not depend on how often a sign was seen during pretraining; it appears to
be driven more by other factors, such as how visually distinctive the sign is
and how much its dictionary production differs from its sentence-context
production.

\FloatBarrier

\subsection{Pseudo-Gloss Pretraining and Translation Check}

Table~\ref{tab:pseudogloss_pretraining_results} isolates pretraining quality
from downstream translation behavior. The stage-1 validation result focuses on pseudo-gloss construction and shows that
rule-based morphological lemmatization is much stronger than a larger raw
surface-token space. The LLM lexical-rule vocabulary is flagged separately because validation F1 is not directly comparable across target vocabularies of different sizes. We include the LLM lexical-rule row for completeness, but interpret F1 primarily within each target vocabulary rather than as a direct fairness comparison across vocabularies of different sizes.

\begin{table}[!h]
\centering
\small
\caption{TSL-News pseudo-gloss pretraining validation. Val F1 is validation
F1-score, the harmonic mean of precision and recall. The gain is measured
against the raw surface-token baseline.}
\label{tab:pseudogloss_pretraining_results}
\begin{tabular}{lrrr}
\toprule
Pseudo-gloss construction & Targets & Val F1 & Gain \\
\midrule
Raw surface tokens & $\sim$16K & 0.1424 & -- \\
LLM lexical-rule & 6539 & 0.4732 & +0.3308 \\
Rule-based morphology lemmas & 4802 & \textbf{0.4791} & \textbf{+0.3367} \\
\bottomrule
\end{tabular}
\end{table}
\FloatBarrier

In the translation check, the Sign2GPT-style translation setup (frozen XGLM decoder with LoRA adapters, as in Fig.~\ref{fig:full_pipeline}) is kept unchanged; only the
encoder initialization differs. We report both LLM lexical-rule and morphology-lemma pretrained encoders; the strongest translation check uses the morphology-lemma pseudo-gloss construction. Reusing the pseudo-gloss pretrained encoder
improves all test bilingual evaluation understudy (BLEU) and recall-oriented
understudy for gisting evaluation (ROUGE)
metrics~\cite{papineni2002bleu,lin2004rouge} over training without stage-1
pretraining. These translation numbers contextualize the changed pretraining
stage rather than introducing a new translation method.

\begin{table}[!h]
\centering
\small
\caption{TSL-News translation check. The translation setup is fixed;
only pseudo-gloss pretraining of the encoder changes. B1--B4 denote
BLEU-1--BLEU-4 and R denotes ROUGE.}
\label{tab:translation_results}
\setlength{\tabcolsep}{4pt}
\begin{tabular}{lrrrrr}
\toprule
Setting & B1 & B2 & B3 & B4 & R \\
\midrule
Without pseudo-gloss pretraining & 23.41 & 16.44 & 12.27 & 9.60 & 23.48 \\
With LLM lexical-rule pretraining & 27.02 & 19.00 & 14.03 & 11.04 & 27.43 \\
With morphology-lemma pretraining & \textbf{29.25} & \textbf{21.06} & \textbf{15.93} & \textbf{12.65} & \textbf{28.87} \\
\bottomrule
\end{tabular}
\end{table}
\FloatBarrier

\subsection{Auxiliary Localization Diagnostics}
\label{Sec:5.3}
Table~\ref{tab:model_scores} compares auxiliary localization diagnostics on
the LLM lexical-rule setup. All columns report temporal IoU except
Top-5@0.50, which reports the percentage of examples whose best top-5
candidate reaches at least 0.50 IoU. Table~\ref{tab:retrieval_recall}
separately reports rank-based retrieval metrics for the all-vocabulary local
NCC setting.

\begin{figure}[!t]
\centering
\includegraphics[width=0.8\linewidth]{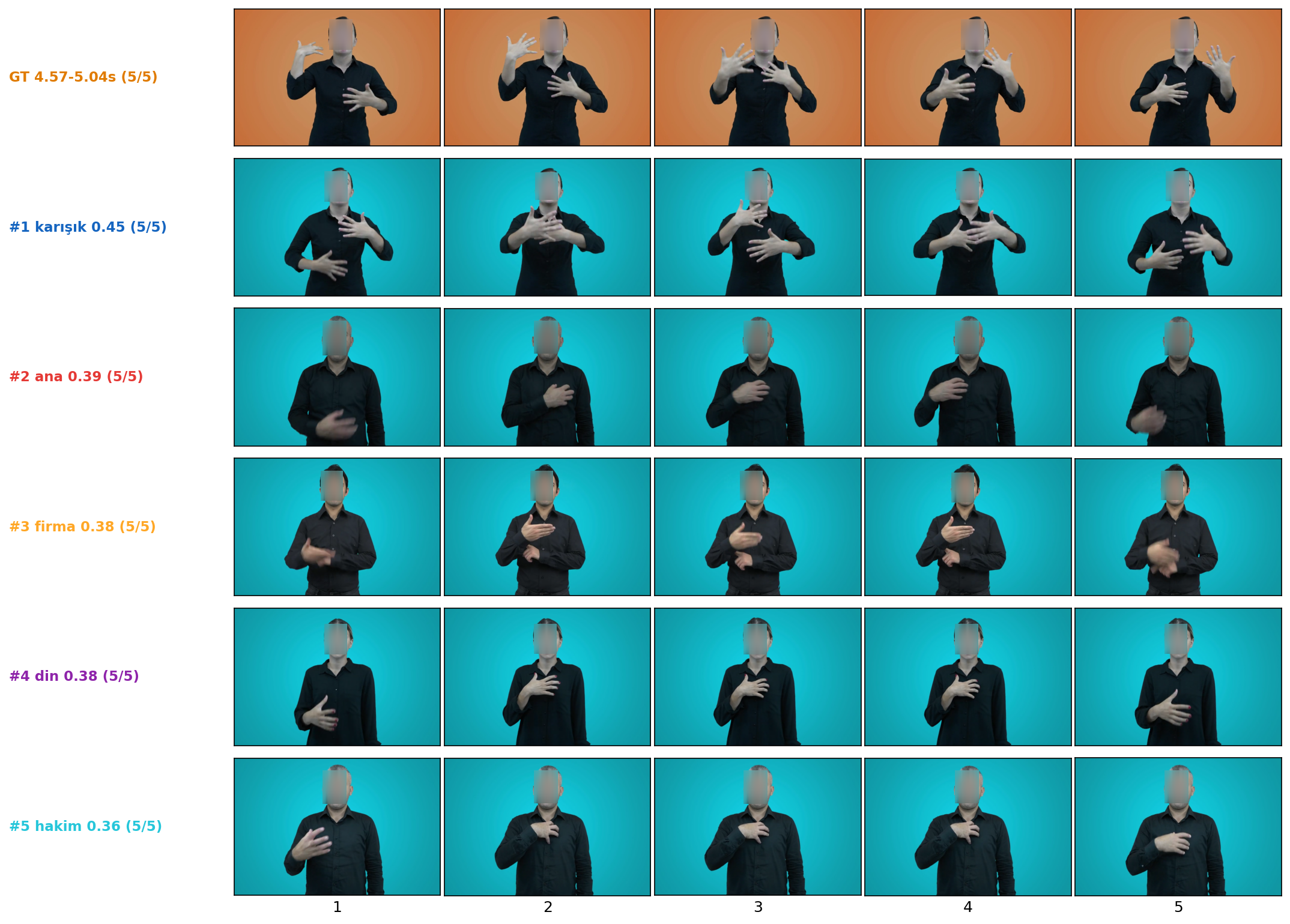}
\caption{Qualitative frame comparison for local all-vocabulary NCC. The top
row shows the ground-truth \emph{Kar{\i}\c{s}{\i}k} (``complicated'') span, and the remaining
rows show the top-5 isolated retrieval candidates. Similar hand and body
configurations support that local NCC retrieves visually nearby sign
representations. Faces are blurred to avoid exposing signer identity.}
\label{fig:local_gt_ncc_frames}
\end{figure}

Figure~\ref{fig:local_gt_ncc_frames} keeps the main qualitative frame evidence by showing frames
from the ground-truth example span and from the isolated videos
of the top-5 retrieved sign-word candidates. If high local NCC scores reflect
visual similarity in the learned sign representations, the top-ranked isolated
signs should show similar hand and body configurations to the ground-truth
span.

\begin{table}[!h]
\centering
\small
\caption{Auxiliary localization diagnostics on the LLM lexical-rule benchmark.
$N$ is the number of evaluated examples for the first two rows and evaluated
segment occurrences for All-vocabulary local NCC.}
\label{tab:model_scores}
\setlength{\tabcolsep}{4pt}
\begin{tabular}{lrrrrr}
\toprule
Diagnostic mode & N & Top-1 & Top-3 & Top-5 & Top-5@0.50 \\
\midrule
Pseudo-gloss classifier & 1817 & 0.241 & 0.279 & 0.280 & 28.9\% \\
Target-known NCC & 1817 & 0.271 & 0.403 & 0.465 & 56.2\% \\
All-vocabulary local NCC & 2071 & 0.377 & 0.482 & 0.516 & 56.3\% \\
\bottomrule
\end{tabular}
\end{table}

\begin{table}[!h]
\centering
\small
\caption{All-vocabulary local NCC retrieval metrics. Recall@$k$ ranks the
correct sign word among all local NCC candidates; Avg. cand. is the mean
number of candidate isolated templates.}
\label{tab:retrieval_recall}
\setlength{\tabcolsep}{3.2pt}
\begin{tabular}{lrrrrrr}
\toprule
Mode & MRR & R@1 & R@5 & R@10 & Med. rank & Avg. cand. \\
\midrule
All-vocabulary local NCC & 0.212 & 13.4\% & 29.1\% & 36.0\% & 33 & 2491 \\
\bottomrule
\end{tabular}
\end{table}

\begin{reviewaddition}
Additional qualitative evaluations for the localization diagnostics are provided in Appendix~\ref{sec:supp_ncc_curve_diagnostics}.
\end{reviewaddition}

\section{Conclusion}

This paper studies whether weak text-derived pseudo-gloss supervision can
pretrain reusable visual representations for limited-resource Turkish Sign
Language. Using TSL-News for broadcast pretraining and TSL-SB for
cross-dataset evaluation, NCC over hidden states shows that the
pretrained encoder transfers
better than raw spatial features, the morphology-lemma pseudo-gloss baseline,
and direct pseudo-gloss score localization. The main conclusion is therefore
about representation learning: even noisy sentence-level pseudo-glosses can
shape a visual encoder into a lexical-temporal representation that supports
dictionary-query sign spotting and provides a useful starting point for
downstream translation checks. The evaluation remains limited to representation
quality through sign spotting: TSL-SB relies on Turkish word-form overlap rather
than expert sign glosses, isolated dictionary productions can differ from
sentence-context productions, TSL-News contains only three signers, and the
all-vocabulary diagnostic still searches within the annotated interval plus a
small buffer. Future work will extend this diagnostic toward
open-ended proposals, sentence-level reranking, broader TSL-SB validation, and
full translation studies that use translation as a consumer of the learned
encoder rather than the central contribution.
The TSL-News corpus and the TSL-SB benchmark annotations will be released upon publication to support research on limited-resource sign languages.

\begin{reviewaddition}
\section*{Acknowledgements}
This work is supported by the Scientific and Technological Research Council of T\"urkiye (T\"UB\.{I}TAK) under the 1001 Scientific and Technological Research Projects Funding Program (Project No.~124E618). We acknowledge the EuroHPC Joint Undertaking for awarding us access to Vega at IZUM, Slovenia, through Development Access allocation 2025D08-090.
\end{reviewaddition}

\bibliographystyle{splncs04}
\bibliography{main}

\clearpage
\appendix
\section{Supplementary Material}
\label{sec:supplementary_material}
\setcounter{figure}{0}
\setcounter{table}{0}
\renewcommand{\thefigure}{\thesection.\arabic{figure}}
\renewcommand{\thetable}{\thesection.\arabic{table}}
\begin{reviewaddition}
\subsection{Sign2GPT Adaptation and Training Configuration}
\label{sec:supp_sign2gpt_comparison}

We retain the Sign2GPT macro-architecture while adapting it for Turkish weak pseudo-gloss pretraining and a controlled translation check. Tables~\ref{tab:supp_stage1_config} and~\ref{tab:supp_stage2_config} specify the frozen and trainable components and the corresponding optimization settings. fastText initializes the prototype vectors but neither selects lemmas nor consolidates Turkish inflections. Because no official trained spaCy-Turkish POS-tagging and lemmatization pipeline was available, pseudo-gloss normalization instead uses the compared Turkish morphological analyzer and constrained LLM mapper.
\end{reviewaddition}

\begin{table}[!htbp]
\centering
\small
\caption{Stage-1 pseudo-gloss pretraining configuration.}
\label{tab:supp_stage1_config}
\setlength{\tabcolsep}{4pt}
\begin{tabular}{p{3.5cm}p{7.4cm}}
\toprule
Component & Value \\
\midrule
Frame backbone & DINOv2 ViT-S/14, frozen base weights \\
Spatial adaptation & LoRA on adaptor layers 9--11; rank 4, dropout 0.1, alpha 4.0 \\
Temporal encoder & MetaFormer, hidden dim 512, heads 8, local attention window 7 \\
Temporal depth & layers [2, 2] with one downsampling stage \\
Prototype head & fastText prototype head, hidden dim 300, dropout 0.2 \\
Temperatures & class/time temperature 0.1, both dynamic \\
Loss & BCE over sentence-level pseudo-gloss set, weight 10.0 \\
Optimizer & AdamW, lr $3{\times}10^{-4}$, weight decay 0.001 \\
Schedule & warmup cosine, 5 warmup epochs, one cycle \\
Batch / accumulation & 4 per GPU / accumulation 2 \\
Augmentation & ImageNet normalization, strength 0.2, random shift 4, stride 2 \\
Max sequence / epochs & 256 frames / 100 epochs \\
Seed and clipping & seed 1, grad norm/value clip 1.0 \\
\bottomrule
\end{tabular}
\end{table}

\begin{table}[!htbp]
\centering
\small
\caption{Stage-2 translation-check configuration.}
\label{tab:supp_stage2_config}
\setlength{\tabcolsep}{4pt}
\begin{tabular}{p{3.5cm}p{7.4cm}}
\toprule
Component & Value \\
\midrule
Language model & \texttt{facebook/xglm-1.7B} \\
Encoder initialization & selected stage-1 checkpoint, then translation training \\
Decoder adaptation & LoRA/adaptor on all 24 layers; rank 4, dropout 0.1, alpha 4.0 \\
Loss & cross entropy with label smoothing 0.1 \\
Optimizer & AdamW, lr $3{\times}10^{-4}$, weight decay 0.001 \\
Schedule & warmup cosine, 5 warmup epochs, one cycle \\
Batch / accumulation & 8 per GPU / accumulation 1 \\
Max epochs & 100 \\
Generation & max length 64, beam size 4, temperature 1.0 \\
Seed and clipping & seed 1, grad norm/value clip 1.0 \\
\bottomrule
\end{tabular}
\end{table}

\FloatBarrier

\subsection{Qualitative NCC Curve Diagnostics}
\label{sec:supp_ncc_curve_diagnostics}

\begin{reviewaddition}
The pseudo-gloss classifier uses a known class-score curve, target-known NCC uses the corresponding isolated template, and all-vocabulary NCC introduces competition among templates. The latter searches only the annotated interval with a 0.25-second margin and is therefore local retrieval rather than full-video open-vocabulary spotting. Its $N$ counts segment occurrences, whereas the other diagnostics count example groups, so their IoU values are not directly comparable.
\end{reviewaddition}

\begin{figure}[!htbp]
\centering
\includegraphics[width=0.9\linewidth]{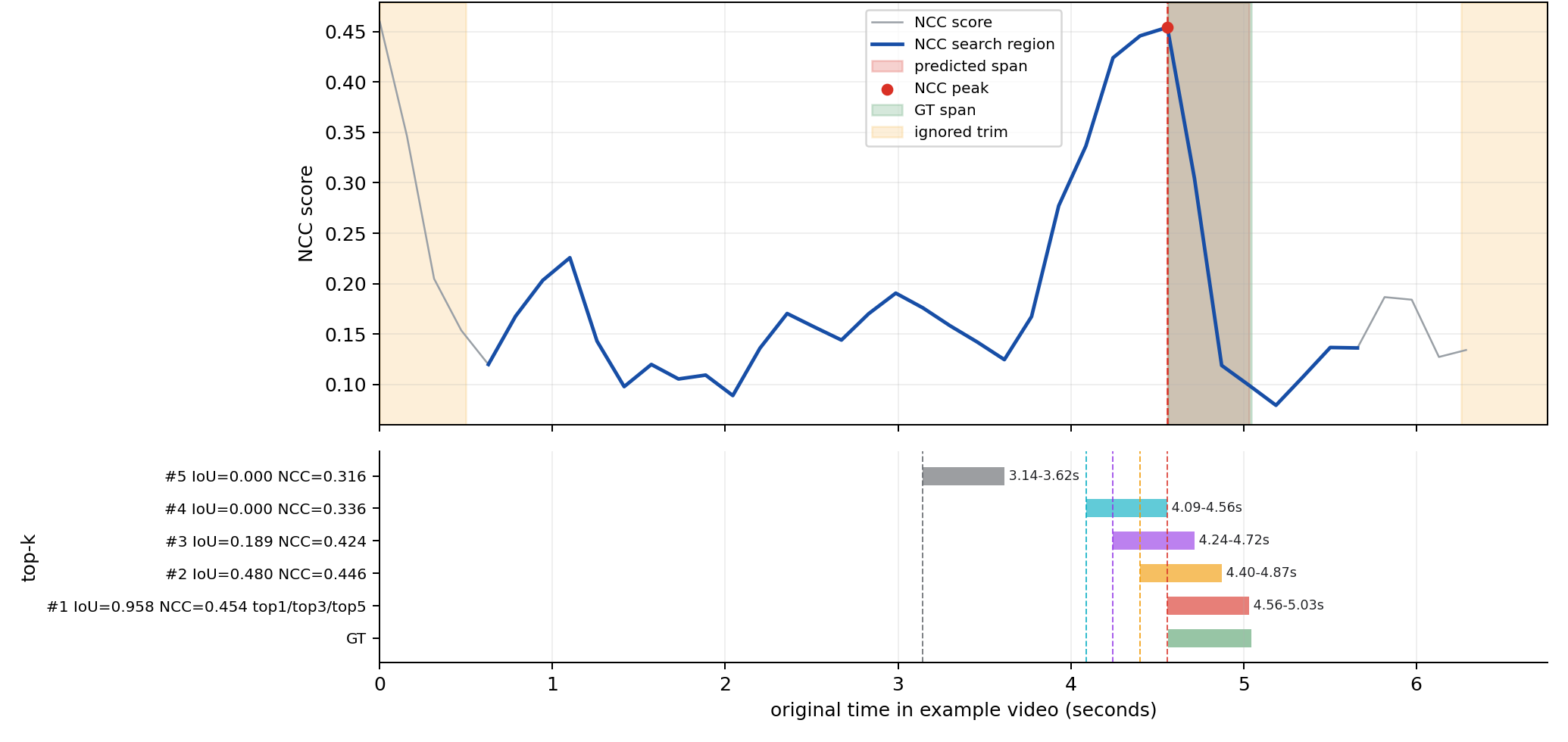}
\caption{Target-known NCC curve diagnostic. The plot visualizes whether the
isolated template for the known target sign produces NCC peaks near the human
annotated sign interval. The searched sign word/gloss in this sequence is
\emph{Kar{\i}\c{s}{\i}k} (``complicated''). The predicted span (pink) largely
overlaps the ground-truth span (green); the span plot below the curve separates
the top-$k$ candidates for easier inspection.}
\label{fig:supp_target_known_ncc_curves}
\end{figure}

\begin{figure}[!htbp]
\centering
\includegraphics[width=0.9\linewidth]{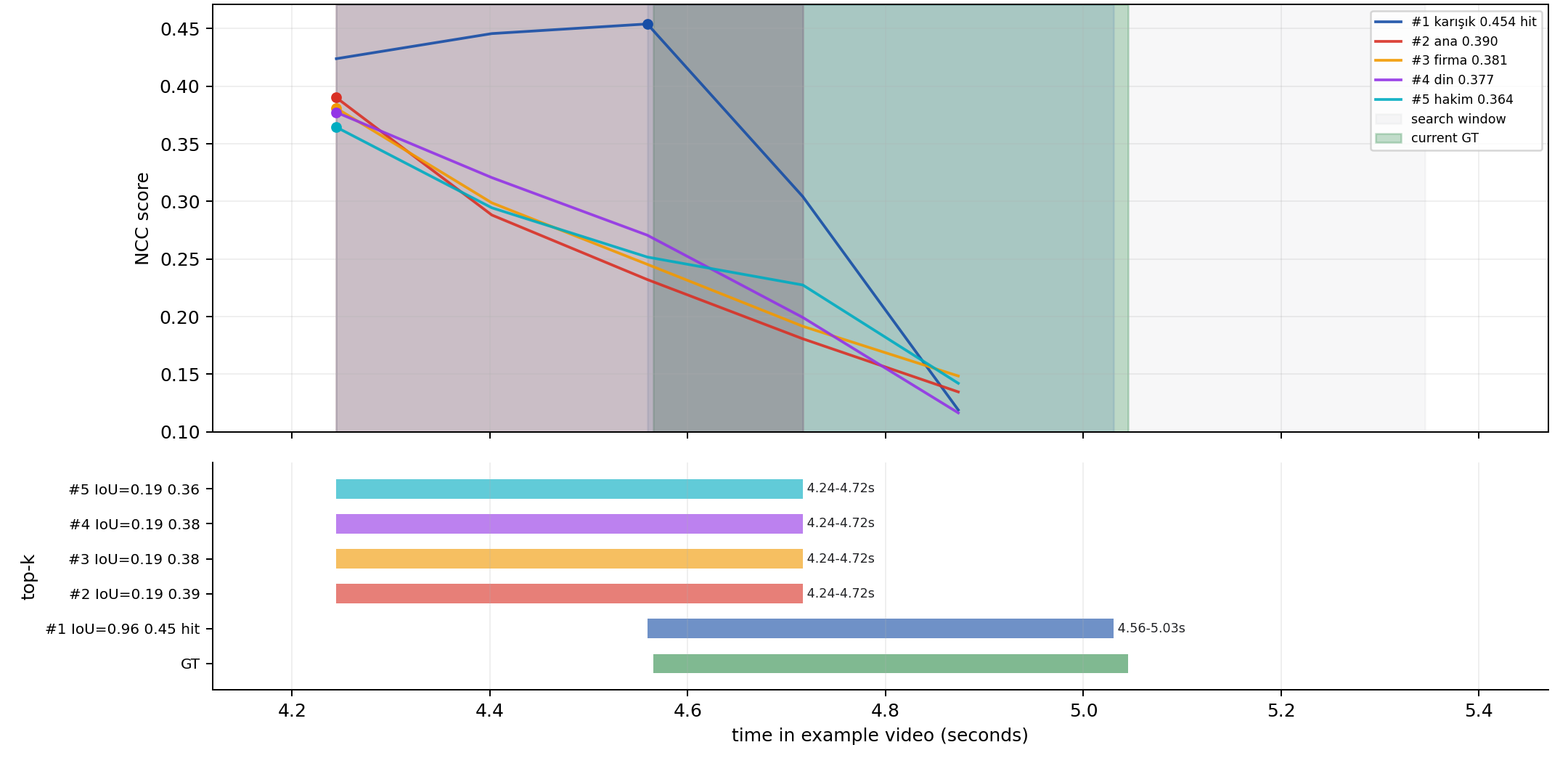}
\caption{Local all-vocabulary NCC diagnostic. The plot visualizes the target
sign against competing TSLD isolated templates inside the annotated interval
plus the 0.25 second buffer. The searched sign word/gloss in this sequence is
\emph{Kar{\i}\c{s}{\i}k} (``complicated''). Each candidate's predicted span is
drawn in a distinct color and largely overlaps the ground-truth span (green);
the span plot below the curve separates the candidates for easier inspection.}
\label{fig:supp_local_gt_ncc_curves}
\end{figure}

Figures~\ref{fig:supp_target_known_ncc_curves}
and~\ref{fig:supp_local_gt_ncc_curves} provide qualitative diagnostics for
the two NCC settings summarized in the main paper. The target-known diagnostic
uses only the isolated template of the annotated target sign and shows whether
its NCC peak aligns with the human temporal annotation. The local
all-vocabulary diagnostic uses the same annotated interval plus a small buffer
but lets the target compete against all candidate isolated templates, making it
visible whether high-ranking alternatives correspond to similar temporal spans
rather than arbitrary background motion.

\FloatBarrier

\subsection{LLM Lexical-Rule Pseudo-Gloss Details}

The LLM lexical-rule pseudo-gloss pipeline is implemented in two stages. First,
candidate pseudo-glosses are constructed from a fixed allowed vocabulary using
exact, root/prefix, and fuzzy matches. Second, a local OpenAI-compatible
chat-completions server selects one candidate or rejects all candidates. The
final experiments use a frozen, manually reviewed merged word-to-pseudo-gloss
mapping; this mapping is then applied deterministically to the TSL-News corpus
CSVs and converted to the stage-1 pseudo-gloss dictionary used in training.

\begin{reviewaddition}
\subsubsection{Worked Pseudo-Gloss Examples}

Table~\ref{tab:supp_pseudogloss_examples} contrasts representative outputs: the rule-based analyzer may return shorter roots such as \emph{kur} and \emph{s\"oyle}, whereas the constrained mapper selects fuller or canonical entries such as \emph{kurulu\c{s}} and \emph{s\"oylemek} from the fixed vocabulary.

\begin{table}[!htbp]
\color{reviewcolor}
\centering
\scriptsize
\caption{Side-by-side pseudo-gloss construction examples from TSL-News. English translations are explanatory only and are not model inputs.}
\label{tab:supp_pseudogloss_examples}
\setlength{\tabcolsep}{3pt}
\begin{tabular}{p{3.3cm}p{3.8cm}p{3.8cm}}
\toprule
Turkish transcript (English) & Morphology-lemma output & LLM lexical-rule output \\
\midrule
\.{I}\c{s}itme engelliler haber b\"ultenine ho\c{s} geldiniz. (Welcome to the news bulletin for the hearing impaired.) & \texttt{i\c{s}it engel haber b\"ulten ho\c{s} gel} & \texttt{i\c{s}itmek engelli haber b\"ulten ho\c{s} gelmek} \\
Cumhurba\c{s}kan{\i} Erdo\u{g}an sanayi kurulu\c{s}lar{\i}nda uygulanan k{\i}smi enerji kesintisinin yak{\i}nda sona erece\u{g}ini s\"oyledi. (The President said the partial energy cut would soon end.) & \texttt{cumhurba\c{s}kan{\i} erdo\u{g}an sanayi kur uygu k{\i}smi enerji kesinti yak son er s\"oyle} & \texttt{cumhurba\c{s}kan{\i} erdo\u{g}an sanayi kurulu\c{s} uygulamak k{\i}smi enerji kesinti yak{\i}nda sona ermek s\"oylemek} \\
\.{I}ki bin dokuz y\"uz yirmi yedi yeni engelli memur atamas{\i}nda sekiz \c{s}ubatta yap{\i}laca\u{g}{\i}n{\i} belirtti. (It stated that 2,927 new disabled civil-servant appointments would be made on 8 February.) & \texttt{iki bin dokuz y\"uz yirmi yedi yen engel memur ata sekiz \c{s}ubat yap belir} & \texttt{iki bin dokuz y\"uz yirmi yedi yeni engelli memur atama sekiz \c{s}ubat yap{\i}lmak belirtmek} \\
\bottomrule
\end{tabular}
\end{table}
\end{reviewaddition}
\FloatBarrier

\begin{table}[!htbp]
\centering
\small
\caption{LLM lexical-rule extraction settings.}
\label{tab:supp_llm_settings}
\setlength{\tabcolsep}{4pt}
\begin{tabular}{p{3.8cm}p{7.1cm}}
\toprule
Item & Setting \\
\midrule
LLM access & local OpenAI-compatible chat-completions server (llama.cpp) \\
Model identifier & \texttt{qwen} local server alias \\
Decoding & temperature 0.0 \\
Word-level batch size & 10 words per LLM request \\
Candidate limit & 15 candidates per source word \\
Timeout / workers & 180 s / 1 worker for word mapping \\
Final mapping entries & 23,162 unique Turkish translation words \\
Non-empty / empty mappings & 23,041 / 121 \\
Exact identity mappings & 4,752 \\
Estimated non-exact LLM requests & at most 1,841 batches with batch size 10 \\
External API cost & none recorded; inference used a local server \\
\bottomrule
\end{tabular}
\end{table}
\FloatBarrier

The word-level system prompt used by the mapper is:

\begin{quote}
\small\ttfamily\raggedright
Turkce kelimeyi pseudo\_gloss ile eslestir.\\
Yalniz ADAYLAR listesinden sec;\\
listede olmayan gloss yazma.\\
Yeni kelime uretme.\\
Kelime cekimli fiil veya ek almis isim olabilir.\\
Kok/anlam olarak dusun. Adaylarda kelimenin kokune\\
benzeyen bir gloss varsa MUTLAKA onu sec.\\
Ornek: gosterdi->gostermek, gosterecek->gostermek,\\
gosteren->gostermek. Ornek: geldiniz->gel,\\
artirdik->artir, cikarilacak->cikar,\\
bultenine->bulten.\\
NONE sadece adaylarin hicbiri kelimeyle anlam veya kok\\
olarak ilgili degilse kullanilir.\\
Sadece JSON dondur:\\
\{"sonuclar":[\{"kelime":"...","gloss":"..."\}]\}
\end{quote}

Its English translation is:

\begin{quote}
\small
Match the Turkish word to a pseudo-gloss. Choose only from the CANDIDATES
list; do not write a gloss that is not in the list, and do not invent a new
word. The word may be an inflected verb or a suffixed noun. Reason by root and
meaning. If one candidate resembles the root of the word, choose it. Use NONE
only when none of the candidates are related to the word by meaning or root.
Return only the requested JSON object.
\end{quote}

The corresponding user prompt has the structure:

\begin{quote}
\small\ttfamily\raggedright
KELIMELER VE ADAYLAR:\\
\{[\\
\quad \{"kelime": word, "candidates":\\
\quad [candidate\_1, ...]\},\\
\quad ...\\
]\}\\
Sadece aday listesinden sec.
\end{quote}

Its English translation is:

\begin{quote}
\small\ttfamily\raggedright
WORDS AND CANDIDATES:\\
\{[\\
\quad \{"word": word, "candidates":\\
\quad [candidate\_1, ...]\},\\
\quad ...\\
]\}\\
Choose only from the candidate list.
\end{quote}

For sentence-level mapping, the alternative mapper uses the same local
chat-completions interface with the default model alias \texttt{qwen},
temperature 0.0, four workers, and candidate chunks of 250. Its system prompt
is JSON-only and closed-vocabulary: select pseudo-glosses only from
\texttt{ADAYLAR}, do not invent new words, and return
\texttt{\{"pseudo\_gloss":["..."]\}}. Repeated identical sentences are cached
within a run, so they receive the same pseudo-gloss set in the generated CSV.
The final reported experiments do not depend on re-calling the LLM: the merged
word mapping is frozen, and rebuilding the 6539-class TSL-News pseudo-gloss
corpus from that mapping is deterministic. A separate independent repeat-call
consistency audit of the original local LLM server was not logged.

\end{document}